\documentclass[letterpaper]{article} 
\usepackage{aaai2027}  
\usepackage[hyphens]{url}  
\usepackage{graphicx} 
\usepackage{natbib}  
\usepackage{caption} 
\usepackage{algorithm}
\usepackage{bm}
\usepackage{algorithmic}
\usepackage{multirow}
\usepackage{amsmath}
\usepackage{amssymb}
\usepackage{makecell}
\usepackage{newfloat}
\usepackage{listings}
\DeclareCaptionStyle{ruled}{labelfont=normalfont,labelsep=colon,strut=off} 
\floatstyle{ruled}
\newfloat{listing}{tb}{lst}{}
\floatname{listing}{Listing}

\usepackage{booktabs}

\title{Learning Where and What to Lift for Bi-planar X-ray-to-CT Reconstruction}
\author{
    Yifei Wu\textsuperscript{\rm 1},
    Yicheng Wu\textsuperscript{\rm 2},
    Qiang Ma\textsuperscript{\rm 3},
    Qi Chen\textsuperscript{\rm 4},
    Renyang Gu\textsuperscript{\rm 5},
    Xinyu Liu\textsuperscript{\rm 6},
    Yongsheng Pan\textsuperscript{\rm 1},
    Yong Xia\textsuperscript{\rm 1}
}

\affiliations{
    \textsuperscript{\rm 1}School of Computer Science, Northwestern Polytechnical University, Xi'an, China\\
    \textsuperscript{\rm 2}Imperial College London, London, UK\\
    \textsuperscript{\rm 3}Department of Brain Sciences, Imperial College London, London, UK\\
    \textsuperscript{\rm 4}Australian Institute for Machine Learning (AIML), Adelaide University, Adelaide, Australia\\
    \textsuperscript{\rm 5}Department of Bioengineering, Imperial College London, London, UK\\
    \textsuperscript{\rm 6}Department of Computing, Imperial College London, London, UK
}

\begin{document}

\maketitle

\begin{abstract}
X-ray imaging can be approximately modeled as the projection of an underlying volumetric attenuation field, with each measurement recording the accumulated attenuation along a corresponding ray path. Reconstructing a CT volume from only a few X-ray views is therefore severely ill-posed, as the projections collapse depth information and leave 3D locations of anatomical regions and their corresponding intensity distributions highly entangled and ambiguous. We observe that once the spatial organization of anatomical regions is established, estimating their CT intensities becomes substantially more tractable.
Motivated by this, we propose \textbf{LiftXR}, an interleaved, geometry-guided framework that explicitly incorporates spatial layout recovery into CT reconstruction. Specifically, a layout lifter first generates a 3D anatomical layout from bi-planar X-rays, providing spatial guidance for an intensity renderer to reconstruct a CT volume. An anatomical parser then performs volumetric perception on the reconstruction, exploiting its spatially resolved boundary and intensity cues to recover a refined anatomical layout. This transition from projection-conditioned layout generation to reconstruction-conditioned anatomical perception allows the parsed layout to provide feedback for region-specific intensity calibration. Extensive experiments on two public datasets demonstrate that LiftXR consistently outperforms recent X-ray-to-CT reconstruction methods, establishing a new state of the art. Moreover, the reconstructed CT achieves superior performance in external downstream segmentation, indicating improved anatomical fidelity. Code will be released.
\end{abstract}

\begin{figure}[t]
\centering
\includegraphics[width=1\columnwidth]{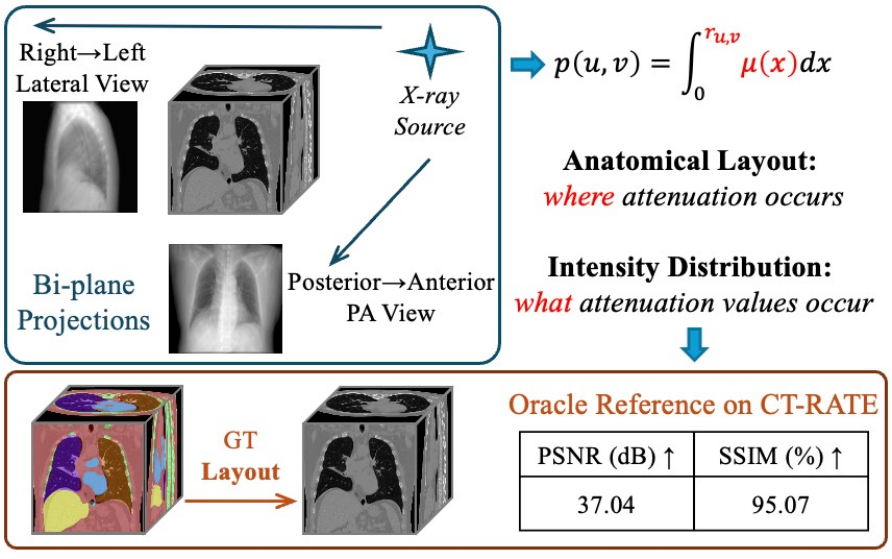}
\caption{{X-ray projection and CT reconstruction}. The projected value $p(u,v)$ records the accumulated attenuation $\mu(\mathbf{x})$ along its corresponding ray $r_{u,v}$. We characterize this process by an anatomical layout capturing the spatial organization of organs and tissues, and an intensity field specifying their voxel-wise CT values. Recovering both aspects from sparse X-ray views is highly ill-posed since geometric structure and intensity information are strongly entangled. Then, an oracle experiment on CT-RATE shows that reconstruction conditioned on ground-truth anatomical layouts achieves 37.04~dB PSNR and 95.07\% SSIM, demonstrating the value of anatomical layout as an effective spatial constraint.
}
\label{fig1}
\end{figure}

\section{Introduction}
Computed tomography (CT) reconstructs volumetric attenuation fields from multi-angle X-ray projections, typically requiring hundreds of views for accurate 3D imaging \cite{jung2021CT}. However, dense-view acquisition increases radiation exposure, limiting its applicability to longitudinal monitoring and large-scale screening \cite{cai2024structure}. Sparse-view CT reconstruction reduces this burden by using only a few dozen projections \cite{lin2023learning,lin2024learning,shi2026dm4ct,guo2024mair}. bi-planar X-ray-to-CT reconstruction offers an even more acquisition-efficient alternative, relying only on frontal and lateral radiographs that are routinely acquired in clinical practice and widely available in public datasets \cite{yu2025spider,wang2017chestx}. Nevertheless, recovering a 3D CT volume from only two projections remains severely ill-posed. 
The ill-posedness originates from the severe loss of depth information during projection acquisition \cite{chen2025mitigating,wu2025dual}.
As illustrated in Fig.~\ref{fig1}, each X-ray pixel records attenuation accumulated along a ray path, collapsing depth-dependent anatomical information into a two-dimensional measurement. We describe the underlying CT volume by an anatomical layout exhibiting the spatial extent and arrangement of organs and tissues, and a volumetric intensity field specifying their voxel-wise CT values. Because distinct 3D layouts and intensity distributions can produce similar 2D projections, the two aspects remain highly entangled under limited observations.
Existing X-ray-to-CT methods \cite{ying2019x2ct,kyung2023perspective,pan2025draw} have introduced geometric consistency, learnable priors, or anatomy-related regularization to facilitate CT reconstruction. However, anatomical organization is typically encoded implicitly within image features or inferred only through the reconstructed intensities, rather than being modeled as an explicit region-level 3D layout. As a result, reconstructed volumes may perform favorably on voxel-wise metrics while still exhibiting inaccurate regional extents, blurred boundaries, or inconsistent spatial relationships \cite{lin2025pixel,shi2025exploration,deo2025metrics}.

To further investigate the role of anatomical layout, we conduct an oracle study on CT-RATE \cite{hamamci2026generalist}, in which the reconstruction model is conditioned on ground-truth anatomical masks. The resulting reconstruction achieves a PSNR of 37.04~dB and an SSIM of 95.07\%. This result demonstrates that the anatomical layout provides strong spatial constraints for CT reconstruction. By localizing major body regions in 3D space, the layout reduces the geometric uncertainty that would otherwise need to be resolved jointly with voxel-wise intensity estimation.

Motivated by these observations, we propose \textbf{LiftXR}, an end-to-end, geometry-guided framework that explicitly incorporates anatomical layout recovery into bi-planar X-ray-to-CT reconstruction. Specifically, a layout lifter first generates an initial 3D anatomical layout from the bi-planar X-rays. This layout provides explicit spatial guidance for an intensity renderer to reconstruct a CT volume. Although imperfect, the reconstructed volume contains spatially resolved boundary and intensity cues, and a subsequent anatomical parser predicts region masks from the reconstructed CT to obtain a refined layout. Finally, an intensity calibrator uses the parsed layout to correct region-specific intensity inconsistencies. Importantly, the two anatomical layouts play fundamentally
different roles. The lifted layout is inferred directly from
bi-planar projections and serves as a generative geometric prior
for volumetric reconstruction. In contrast, the parsed layout is
obtained by perceiving anatomical regions from the initially
reconstructed CT volume, constituting an explicit volumetric
anatomical parsing task. This transition from layout generation
to layout perception enables the reconstructed volume to provide
feedback for geometry refinement, which subsequently guides
region-specific intensity calibration.

Our contributions are threefold:

\begin{itemize}
\item We formulate the bi-planar X-ray-to-CT reconstruction task by explicitly modeling anatomical layout as a spatial constraint for volumetric generation. By specifying \textit{where} anatomical regions are located in 3D space, the layout constrains and guides the estimation of \textit{what} CT intensities they exhibit.

\item We propose \textbf{LiftXR}, a geometry-guided framework that couples volumetric generation with anatomical perception. LiftXR integrates projection-conditioned layout generation, conditional CT rendering, reconstruction-conditioned anatomical parsing, and region-specific intensity calibration, forming a unified generation-perception feedback process.

\item Extensive experiments on two public chest CT datasets demonstrate that LiftXR achieves state-of-the-art reconstruction performance. Its reconstructed CT volumes further yield superior downstream segmentation results, showcasing improved anatomical fidelity.
\end{itemize}

\section{Related Work}
\subsection{X-ray-to-CT Reconstruction}
X-ray-to-CT reconstruction aims to recover a 3D CT volume from one or two 2D X-ray projections, which is highly challenging due to the severe ambiguity introduced by the projection process.
X2CT \cite{ying2019x2ct} first demonstrated the feasibility of reconstructing volumetric CT from bi-planar X-rays, and subsequent studies introduced geometric, generative, and structural priors to reduce this ambiguity.
PerX2CT \cite{kyung2023perspective} incorporates perspective projection modeling to improve spatial localization, DiffuX2CT \cite{liu2024diffux2ct} introduces a conditional diffusion prior to improve volumetric consistency, and DSDF \cite{pan2025draw} separates global structure modeling from local texture synthesis to enhance reconstruction quality.
However, these methods mainly optimize voxel-level appearance or represent anatomical structure implicitly, leaving explicit constraints on organ boundaries, spatial extents, and inter-organ relationships underexplored \cite{lin2025pixel}.
LiftXR addresses this limitation by explicitly predicting anatomical structures, where predicted layouts guide CT estimation and reconstructed volumes provide cues for further layout parsing.
This interleaved layout-intensity interaction constrains the solution space and promotes anatomically consistent reconstruction with clear anatomical constraints derived from bi-planar X-rays.

\subsection{Sparse-view CT Reconstruction}
Sparse-view CT reconstruction aims to reduce radiation exposure by recovering CT volumes from limited X-ray projections.
Analytical methods such as FBP and FDK require sufficient angular sampling and often introduce streak artifacts and structural distortions under severe undersampling \cite{feldkamp1984FDK}.
Traditional approaches \cite{yang2025ct,niu2014sparse} alleviate this ill-posedness with handcrafted regularization priors, but they typically require expensive optimization and careful parameter tuning.
Deep learning methods improve reconstruction by learning effective data-driven priors from large-scale datasets \cite{chen2018learn,lin2019dudonet}.
Image-domain approaches, such as FBPConvNet \cite{jin2017deep}, refine degraded analytical reconstructions, whereas projection-domain approaches recover missing measurements before analytical reconstruction through DNN-based sinogram synthesis \cite{lee2018deep} or learned interpolation \cite{dong2019sinogram}. 
Nevertheless, these approaches still depend on multiple projections collected along a scanning trajectory.
In contrast, X-ray-to-CT reconstruction relies on only one or two radiographs, leading to a more severely under-constrained inverse problem that requires stronger anatomical priors.

\begin{figure*}[!htb]
\centering
\includegraphics[width=1.0\textwidth]{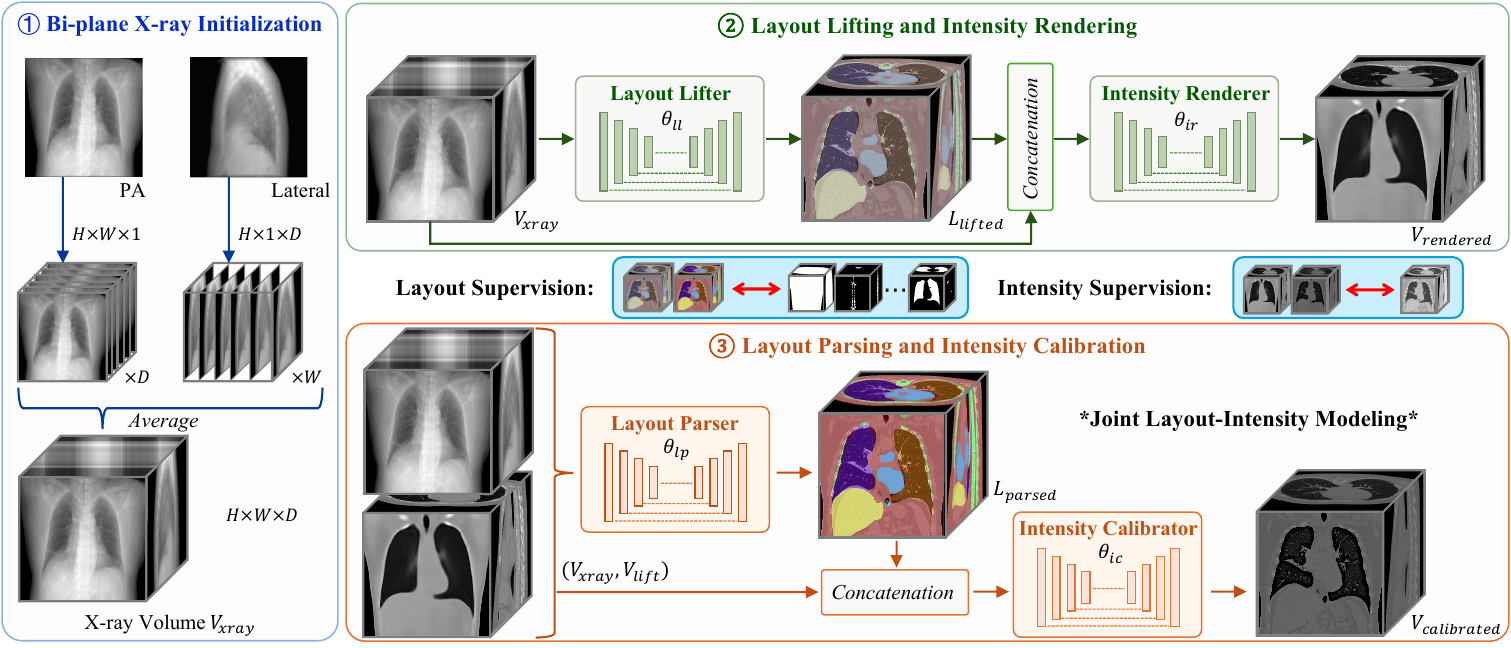}
\caption{{Overview of LiftXR}, an interleaved, geometry-guided framework coupling volumetric generation with anatomical perception. A Layout Lifter first performs projection-conditioned anatomical generation, which guides an Intensity Renderer to reconstruct an initial CT volume. A Layout Parser then performs reconstruction-conditioned anatomical perception using the spatially resolved boundary and intensity cues. The parsed layout is subsequently fed back to an Intensity Calibrator to correct region-specific intensity distributions.}
\label{fig_overall}
\end{figure*}

\subsection{Medical Image Generation}
Generative models, particularly CNN-GANs and diffusion models, have been widely applied to medical image reconstruction \cite{song2026learning,lin2026real,pan2026x2shape}.
CNN-GANs learn supervised input--target mappings through convolutional generators, while adversarial discriminators promote realistic textures and plausible structures \cite{goodfellow2014generative,isola2017image}.
They have been applied to accelerated MRI, low-dose CT, and cross-modality translation \cite{yang2017dagan,liu2023solving,armanious2020medgan,Wu_2024_CVPR,wu2026codebrain}, offering efficient inference and structural control.
Diffusion models generate high-quality images through iterative denoising \cite{ho2020denoising} and have been extended to MRI, sparse-view CT, and low-dose CT reconstruction \cite{webber2024diffusion,chung2022score}.
Despite their strong generative capability, diffusion models require multi-step sampling, resulting in considerable computational overhead for 3D reconstruction and structural uncertainty \cite{jiang2025fast,chen2025cross}.
Therefore, we adopt a CNN-GAN framework, which provides efficient deterministic inference and better integrates with anatomy-conditioned CT reconstruction.

\section{Method}
LiftXR reconstructs a CT volume from paired PA and lateral X-rays by interleaving anatomical layout recovery with volumetric intensity estimation. As illustrated in Fig.~\ref{fig_overall}, the two projections are first up-sampled and fused into an explicit 3D X-ray volume. A Layout Lifter predicts a coarse anatomical layout, which constrains an Intensity Renderer to generate an initial CT volume. Although imperfect, this reconstruction provides spatially resolved boundary and intensity cues for a Layout Parser to refine the anatomical layout, further regularizing regional occupancy and spatial relationships. The refined layout is then combined with the volumetric representation to guide an Intensity Calibrator in correcting region-specific intensity inconsistencies. Through this reciprocal interaction between layout and intensity, LiftXR progressively resolves geometric ambiguity and improves anatomical fidelity.

\noindent\subsection{Bi-planar X-ray Initialization}
Given two PA and LAT projections, we first construct an explicit X-ray volume to bridge the gap between two-dimensional observations and the target three-dimensional space.
Each projection is replicated along its unobserved spatial dimension and aligned within a shared volumetric coordinate system for subsequent aggregation.
\begin{equation}
V_{\mathrm{PA}}=\operatorname{Replicate}(X_{\mathrm{PA}},D), 
V_{\mathrm{LAT}}=\operatorname{Replicate}(X_{\mathrm{LAT}},W),
\end{equation}
where $V_{\mathrm{PA}},V_{\mathrm{LAT}}\in\mathbb{R}^{H\times W\times D}$ denote the lifted representations of the two views. 
We then aggregate these representations to obtain the X-ray volume by voxel-wise averaging
\begin{equation}
V_{\mathrm{xray}}=\operatorname{Average}
\left(V_{\mathrm{PA}},V_{\mathrm{LAT}}\right),
\end{equation}
which preserves the ray-integrated observations from both projections in a unified representation.
This strategy can naturally extend to additional projection angles by lifting each view into the same volumetric coordinate system and applying the same aggregation operation.
However, a direct replication operation cannot recover the missing depth information or resolve the inherent projection ambiguity.

\subsection{Layout Lifting and Intensity Rendering}
Therefore, starting from the X-ray volume $V_{\mathrm{xray}}$, LiftXR performs geometry-guided volumetric lifting to establish an initial volumetric state.
Instead of directly predicting intensity from sparse projections, a Layout Lifter $\theta_{ll}$ first estimates a 3D anatomical layout that provides explicit geometric constraints for subsequent CT reconstruction.
The layout is predicted solely based on $V_{\mathrm{xray}}$ as
\begin{equation}
L_{\mathrm{lifted}}
=
\theta_{ll}
(V_{\mathrm{xray}}),
\end{equation}
where $L_{\mathrm{lifted}}$ represents the lifted layout.
By encoding organ occupancy, spatial extent, and relative arrangement, $L_{\mathrm{lifted}}$ provides anatomical configurations compatible with the bi-planar observations.
Conditioned on the projection-derived representation and predicted layout, the Intensity Renderer $\theta_{ir}$ reconstructs the coarse CT volume as
\begin{equation}
V_{\mathrm{rendered}}
=
\theta_{ir}
\left(
V_{\mathrm{xray}}
\oplus
L_{\mathrm{lifted}}
\right)
\end{equation}
where $\oplus$ is the channel-wise concatenation. This layout-conditioned lifting process transforms the X-ray volume into a structured volumetric state that provides a geometry-aware initialization for subsequent refinement.

\subsection{Layout Parsing and Intensity Calibration}
The lifted layout and CT volume remain imperfect due to limited information from bi-planar projections.
LiftXR therefore introduces an interleaved refinement stage that parses the anatomical layout and calibrates the CT intensity field.
\begin{equation}
L_{\mathrm{parsed}}
=
\theta_{lp}
\left(
V_{\mathrm{xray}}
\oplus
V_{\mathrm{rendered}}
\right),
\end{equation}
where $V_{\mathrm{xray}}$ preserves projection evidence and
$V_{\mathrm{rendered}}$ provides spatially resolved boundary and attenuation cues for the layout refinement by a parser $\theta_{lp}$.
Conditioned on the refined layout $L_{\mathrm{parsed}}$, the Intensity Calibrator $\theta_{ic}$ further updates the rendered CT volume as
\begin{equation}
V_{\mathrm{calibrated}}
=
\theta_{ic}
\left(
V_{\mathrm{xray}}
\oplus
V_{\mathrm{rendered}}
\oplus
L_{\mathrm{parsed}}
\right).
\end{equation}

Overall, although both stages predict anatomical layouts, they solve different tasks. The Layout Lifter performs projection-conditioned layout generation to initialize the missing 3D geometry, whereas the Layout Parser performs reconstruction-conditioned anatomical perception to recover spatially resolved organ boundaries and regional identities.

\subsection{Interleaved Layout-Intensity Supervision}
There are two aspects to supervise the anatomical layout prediction and CT reconstruction.
We denote predicted layouts as $\mathcal{A}=\{L_{\mathrm{lifted}},L_{\mathrm{parsed}}\}$ and reconstructed volumes as $\mathcal{V}=\{V_{\mathrm{rendered}},V_{\mathrm{calibrated}}\}$.
Layout supervision constrains 3D anatomical geometry, while CT supervision preserves intensity and volumetric appearance.

\subsubsection{Anatomical Layout Supervision:}
Accurate anatomical layouts provide explicit constraints on organ occupancy, boundary placement, and spatial organization.
We extract seven parenchymal and major thoracoabdominal regions, including {tissue}, {bone}, {heart}, {liver}, {left lung}, {right lung}, and {esophagus}, for spatial localization.
The target anatomical layout is extracted as
\begin{equation}
L_{\mathrm{GT}}
=
E_{\mathrm{seg}}
\left(
V_{\mathrm{GT}}
\right),
\end{equation}
where $E_{\mathrm{seg}}(\cdot)$ denotes the anatomical mask extractor.
Both the lifted layout and parsed layout are supervised by
\begin{equation}
\mathcal{L}_{\mathrm{layout}}
=
\sum_{a\in\mathcal{A}}
\left[
\operatorname{CE}
\left(
a,L_{\mathrm{GT}}
\right)
+
\operatorname{Dice}
\left(
a,L_{\mathrm{GT}}
\right)
\right],
\end{equation}
where $a$ is selected from $\mathcal{A}$.
This supervision guides both initial layout lifting and subsequent layout parsing toward anatomically plausible three-dimensional structures.

\subsubsection{CT Reconstruction Supervision:}
While anatomical layouts provide explicit geometric constraints, they cannot fully determine region-specific intensity distributions.
Therefore, both rendered and calibrated CT volumes are supervised using the ground-truth CT volume.
Following \cite{ying2019x2ct}, for each reconstructed volume $v\in\mathcal{V}$, the multi-scale feature matching loss is defined as
\begin{equation}
\mathcal{L}_{\mathrm{MSL}}^{(v)}
=
\left\|
MS(v)
-
\mathrm{sg}
\left(
MS(V_{\mathrm{GT}})
\right)
\right\|_1,
\end{equation}
where $MS(\cdot)$ denotes the multi-scale feature transformation, and $\mathrm{sg}(\cdot)$ denotes the stop-gradient operation.
The CT reconstruction objective is defined as
\begin{equation}
\mathcal{L}_{\mathrm{rec}}
=
\sum_{v\in\mathcal{V}}
\left[
\mathcal{L}_{1}
\left(
v,V_{\mathrm{GT}}
\right)
+
\mathcal{L}_{\mathrm{MSL}}^{(v)}
+
\mathcal{L}_{\mathrm{adv}}^{(v)}
\right],
\end{equation}
where $\mathcal{L}_1$ is the mean absolute error (MAE) loss and $\mathcal{L}_{\mathrm{adv}}^{(v)}$ denotes the adversarial loss that encourages reconstructed CT volumes to match the distribution of real CT volumes, following the typical setting \cite{ying2019x2ct,pan2025draw}.
This objective preserves voxel-level intensities, multi-scale structural consistency, and realistic volumetric appearance throughout the lifting and calibration stages.

\subsubsection{Overall Training Objective:}
The anatomical layout and CT reconstruction objectives jointly optimize the shape and intensity branches of LiftXR.
The overall training objective is defined as
\begin{equation}
\mathcal{L}_{\mathrm{total}}
=
\mathcal{L}_{\mathrm{rec}}
+
\lambda \times
\mathcal{L}_{\mathrm{layout}}
,
\end{equation}
where $\lambda$ is a loss weight to balance the two aspects.

\begin{table*}[t]
\centering
\setlength{\tabcolsep}{4.5pt}
\renewcommand{\arraystretch}{1.1}
\scriptsize
\resizebox{\textwidth}{!}{%
\begin{tabular}{l cccc cccc}
\toprule[1.25pt]
\multirow{3}{*}{Method}
& \multicolumn{4}{c}{CT-RATE}
& \multicolumn{4}{c}{LIDC-IDRI} \\
\cmidrule(lr){2-5}
\cmidrule(lr){6-9}
& \multicolumn{2}{c}{Reconstruction}
& \multicolumn{2}{c}{Segmentation}
& \multicolumn{2}{c}{Reconstruction}
& \multicolumn{2}{c}{Segmentation} \\
\cmidrule(lr){2-3}
\cmidrule(lr){4-5}
\cmidrule(lr){6-7}
\cmidrule(lr){8-9}
& PSNR (dB)$\uparrow$
& SSIM (\%)$\uparrow$
& Dice (\%)$\uparrow$
& HD95 (voxel)$\downarrow$
& PSNR (dB)$\uparrow$
& SSIM (\%)$\uparrow$
& Dice (\%)$\uparrow$
& HD95 (voxel)$\downarrow$ \\
\midrule

X2CT \cite{ying2019x2ct}  
& $25.96_{\pm 1.00}$
& $71.88_{\pm 0.51}$
& $63.39_{\pm 0.76}$
& $16.05_{\pm 2.12}$
& $25.06_{\pm 1.13}$
& $69.47_{\pm 0.57}$
& $56.80_{\pm 0.73}$
& $19.81_{\pm 1.45}$ \\

PerX2CT \cite{kyung2023perspective}    
& $25.84_{\pm 0.84}$
& $71.29_{\pm 0.36}$
& $55.44_{\pm 0.59}$
& $19.19_{\pm 1.64}$
& $25.34_{\pm 0.87}$
& {$\underline{70.24}_{\pm 0.60}$}
& $58.96_{\pm 0.78}$
& $18.23_{\pm 1.03}$ \\

DSDF \cite{pan2025draw}  
& $25.82_{\pm 0.99}$
& {$\underline{73.29}_{\pm 0.39}$}
& {$\underline{66.09}_{\pm 0.79}$}
& {$\underline{13.92}_{\pm 1.81}$}
& {$\underline{25.73}_{\pm 0.94}$}
& $69.43_{\pm 0.58}$
& $58.38_{\pm 0.89}$
& {$\underline{16.87}_{\pm 1.01}$} \\
\midrule

GAAL \cite{liu2024geometry}     
& $22.82_{\pm 1.21}$
& $65.62_{\pm 0.40}$
& $28.21_{\pm 0.66}$
& $94.19_{\pm 3.50}$
& $21.02_{\pm 1.23}$
& $51.53_{\pm 0.52}$
& $30.09_{\pm 0.98}$
& $85.56_{\pm 4.98}$ \\

DX2CT \cite{jeong2025dx2ct}  
& $25.76_{\pm 0.49}$
& $70.42_{\pm 0.15}$
& $64.11_{\pm 0.49}$
& $14.98_{\pm 1.23}$
& $25.06_{\pm 1.34}$
& $69.27_{\pm 0.44}$
& $59.02_{\pm 1.09}$
& $17.45_{\pm 1.49}$ \\

DiffuX2CT \cite{liu2024diffux2ct}  
& {$\underline{26.03}_{\pm 0.60}$}
& $72.82_{\pm 0.19}$
& ${65.58}_{\pm 0.51}$
& $14.82_{\pm 1.12}$
& $25.12_{\pm 1.35}$
& $70.16_{\pm 0.40}$
& {$\underline{59.22}_{\pm {0.95}}$}
& ${17.07}_{\pm 1.40}$ \\
\midrule

\textbf{LiftXR} (Ours)      
& ${{\mathbf{26.45}_{\pm 1.00}}}^{**}$
& ${\mathbf{76.44}_{\pm 0.40}}^{**}$
& ${\mathbf{68.64}_{\pm 0.80}}^{**}$
& ${\mathbf{12.51}_{\pm 1.60}}^{**}$
& ${\mathbf{26.03}_{\pm 1.08}}^{*}$
& {${\mathbf{71.72}_{\pm 0.60}}^{**}$}
& {${\mathbf{62.16}_{\pm 0.87}}^{**}$}
& ${\mathbf{15.91}_{\pm 1.14}}^{**}$ \\

\bottomrule[1.25pt]
\end{tabular}%
}
\caption{{Quantitative comparisons} on the CT-RATE and LIDC-IDRI datasets. Reconstruction and segmentation results are reported as mean $\pm$ standard deviation. The {best} and {second-best} results are highlighted in bold and underlined, respectively. $^{*}$ and $^{**}$ indicate that LiftXR significantly outperforms the second-best method at $p < 0.05$ and $p < 0.01$, respectively.}
\label{tab:quantitative}
\end{table*}

\section{Experiment and Result}
\subsection{Implementation Details}
\paragraph{Dataset and Pre-processing:}
We evaluate LiftXR on two benchmarking chest CT datasets. 
On the large-scale CT-RATE dataset \cite{hamamci2026generalist}, we randomly select 2,000 subjects for training and 300 for testing.
On the LIDC-IDRI dataset, which contains 1,018 CT volumes \cite{armato2011lung}, we use 818 volumes for training and 200 for testing.
Following the official setting, anatomical labels are generated on both datasets by using a frozen TotalSegmentator model \cite{wasserthal2023totalsegmentator}.
Following prior X-ray-to-CT studies \cite{ying2019x2ct,gopalakrishnan2022fast}, we generate PA and Lateral radiographs from each CT volume using digitally reconstructed radiography (DRR) techniques, yielding synthetic pairs. 
We use the DiffDRR \cite{gopalakrishnan2022fast} tool\footnote{\url{https://github.com/eigenvivek/DiffDRR}}, with the setting using a source-to-detector distance of 3,000 mm and a detector pixel spacing of 1.0 mm.
All CT volumes are resized to $128 \times 128 \times 128$. The intensity range is clamped to $[-1024,1024]$ HU and then normalized to $[0,1]$.

\paragraph{Experimental Settings:}
We train LiftXR using the Adam optimizer with a learning rate of $2e^{-4}$
and a batch size of 4. 
Based on the sensitivity analysis, $\lambda$ is set to 1.
$E_{\mathrm{seg}}(\cdot)$ is set to TotalSegmentator to obtain the anchor organs.
100,000 iterations are used for training, taking approximately 60 hours on a single NVIDIA GeForce RTX 4090 GPU.  
Each model inside our LiftXR is a 3D U-Net architecture, and the total computational complexity is 6480 GFLOPs with 172.62M parameters.
To ensure fair comparisons, all experiments use a fixed random seed 43 and follow identical settings. For external downstream segmentation, we mainly focus on thoracoabdominal regions, and 20 specific categories are selected, which are elaborated in the supplementary material.

\subsection{Main Results}
\paragraph{Quantitative Analysis:}
Tab.~\ref{tab:quantitative} compares LiftXR with representative X-ray-to-CT reconstruction methods on CT-RATE and LIDC-IDRI.
LiftXR achieves the best performance across all reconstruction and segmentation metrics on both datasets, demonstrating consistent effectiveness in recovering CT intensities and anatomically meaningful three-dimensional structures.
The improvements in PSNR are relatively moderate compared with those in anatomical metrics, as voxel-wise errors are dominated by large homogeneous regions and are less sensitive to localized corrections around organ boundaries and small structures.
In contrast, the consistent SSIM improvements indicate that LiftXR better preserves local structural patterns and tissue contrast during reconstruction.
These results suggest that explicit anatomical layout modeling reduces the depth ambiguity inherent in bi-planar projections and constrains the reconstruction toward anatomically plausible volumetric configurations.
These gains indicate that the predicted layout explicitly constrains organ extent, shape, and spatial relationships, preventing anatomically implausible solutions that may still achieve comparable voxel-wise similarity.
Meanwhile, the reconstructed CT volume provides boundary and tissue-specific intensity cues to refine uncertain layouts.
The refined layout subsequently guides CT calibration to reduce tissue mixing, geometric distortions, and inconsistent intensity distributions.
Through this interleaved refinement process, LiftXR progressively improves both organ occupancy and boundary accuracy.

\begin{figure*}[!htb]
    \centering
    \includegraphics[width=1.0\textwidth]{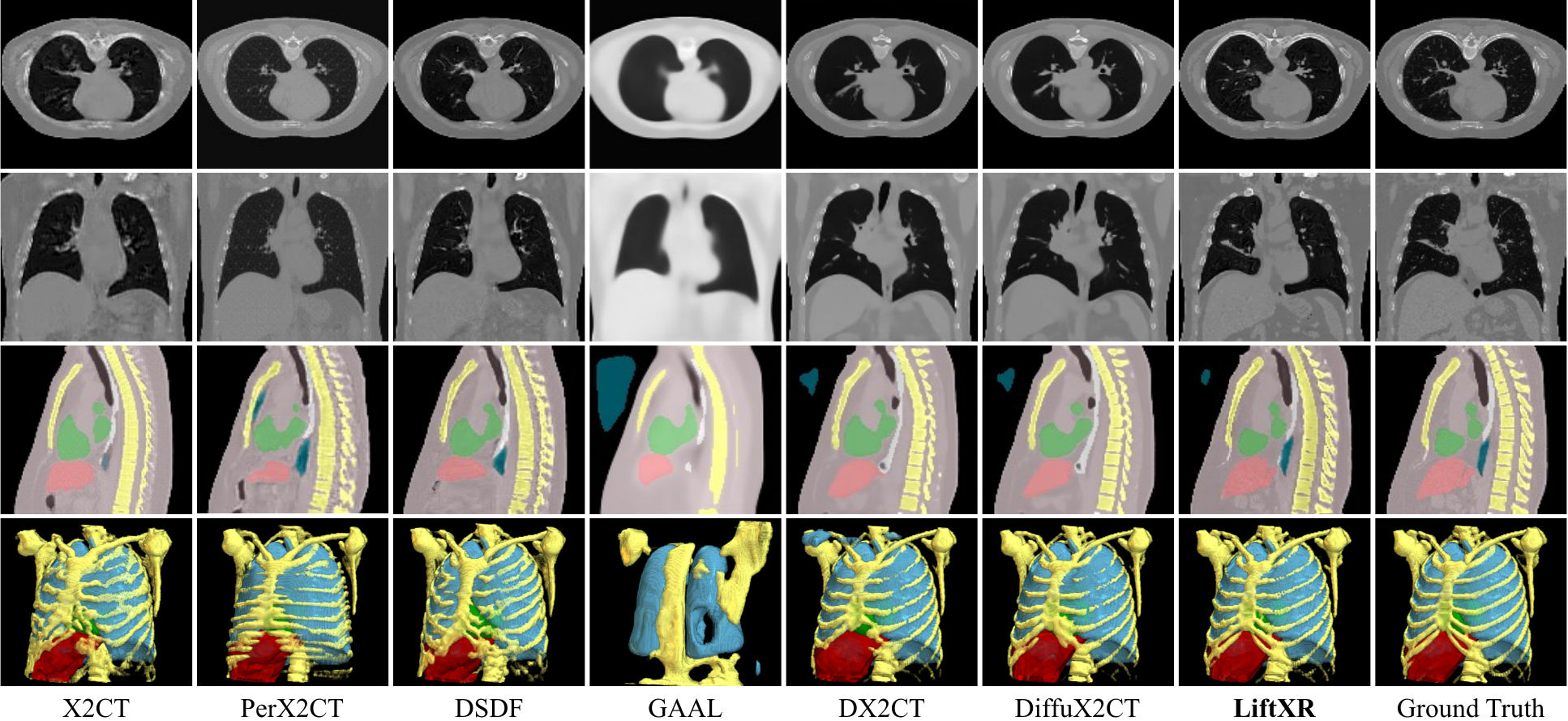} 
    \caption{{Qualitative comparisons} on the CT-RATE dataset. The top two rows show axial and coronal slices, while the bottom two rows present the corresponding segmentation layouts in 2D and 3D. LiftXR more faithfully preserves anatomical structures and recovers fine-grained details, such as maintaining rib continuity and reconstructing a smooth liver contour.}
    \label{fig2}
\end{figure*}

\paragraph{Qualitative Analysis:}
Fig.~\ref{fig2} compares axial and coronal CT slices, sagittal segmentation overlays, and 3D anatomical renderings on the CT-RATE dataset.
Existing methods generally recover the overall thoracic structure but exhibit noticeable errors in both slice-level details and volumetric anatomy.
X2CT, PerX2CT, and DSDF preserve the general lung regions but produce blurred mediastinal boundaries, missing fine structures, and inaccurate organ shapes.
GAAL, a multi-view reconstruction method, exhibits severe over-smoothing and substantial loss of anatomical details, likely due to the difficulty of inferring fine structures from limited projection constraints. 
DX2CT and DiffuX2CT, diffusion-based approaches, recover cleaner global structures with improved volumetric coherence, but still introduce local boundary errors and inconsistencies in inter-organ relationships.
In comparison, LiftXR produces results closer to the ground truth, with clearer lung contours and more accurate vertebral and rib structures.
The final row removes surrounding tissues to visualize internal anatomy, where LiftXR maintains more continuous ribs, better spine alignment, more symmetric lungs, and more plausible organ arrangements.
These localized geometric improvements explain the moderate PSNR gains and the larger improvements in anatomical metrics, which are more sensitive to organ extent and boundary accuracy.
The qualitative observations agree with the quantitative results, demonstrating that interleaved refinement of anatomical layout and CT intensity enables more accurate volumetric reconstruction.

\begin{table}[t]
\centering
\setlength{\tabcolsep}{6pt}
\renewcommand{\arraystretch}{1.1}
\scriptsize
\resizebox{\columnwidth}{!}{%
\begin{tabular}{cccccc}
\toprule[1.25pt]
\multirow{2}{*}{Layout} 
& \multirow{2}{*}{\makecell[c]{Interleaved\\Training}}
& \multicolumn{2}{c}{Reconstruction}
& \multicolumn{2}{c}{Segmentation} \\
\cmidrule(lr){3-4}
\cmidrule(lr){5-6}
& 

& PSNR (dB)$\uparrow$
& SSIM (\%)$\uparrow$
& Dice (\%)$\uparrow$
& HD95 (voxel)$\downarrow$
\\
\midrule
- 
& - 
& 25.68 
& 72.89
&  63.54
& 15.45
 \\ 

& $\checkmark$ 
& \underline{26.07}
& \underline{75.34}
& \underline{67.03}
& \underline{{13.31}}
 \\ 

$\checkmark$  
& - 
& 25.95
& 74.10
& 65.45
& 13.45
 \\ 

$\checkmark$ 
& $\checkmark$ 
& \textbf{{26.45}} 
& \textbf{{76.44}}
& \textbf{68.64}
& \textbf{12.51} \\ 
\bottomrule[1.25pt]
\end{tabular}%
}
\caption{{Ablation studies} on the CT-RATE dataset.
{Best} and {second-best} results are highlighted. Both anatomical layout and interleaved training are evaluated for both reconstruction and downstream segmentation.
}
\label{tab:ablation}
\end{table}

\paragraph{Ablation Study:}
Tab.~\ref{tab:ablation} reports the ablation results on CT-RATE.
Both the layout and interleaved training contribute to reconstruction and segmentation performance, while their combination provides complementary benefits.
Anchor organs establish explicit organ-level references, constraining the possible anatomical configurations during the initial lifting process and reducing large deviations in organ extent and spatial arrangement.
The interleaved training further improves the reconstruction by enabling interleaved interaction between the anatomical layout and CT volume, where the estimated geometry and intensity information mutually correct residual structural and attenuation errors.
While anchor organs mainly provide global anatomical guidance, the interleaved training focuses on improving local details, including organ boundaries, tissue appearance, and inter-organ relationships.
Their effects demonstrate that reliable anatomical prediction together with interleaved layout-intensity refinement is effective in X-ray-to-CT reconstruction.

\begin{figure}[t]
    \centering
    \includegraphics[width=1.0\columnwidth]{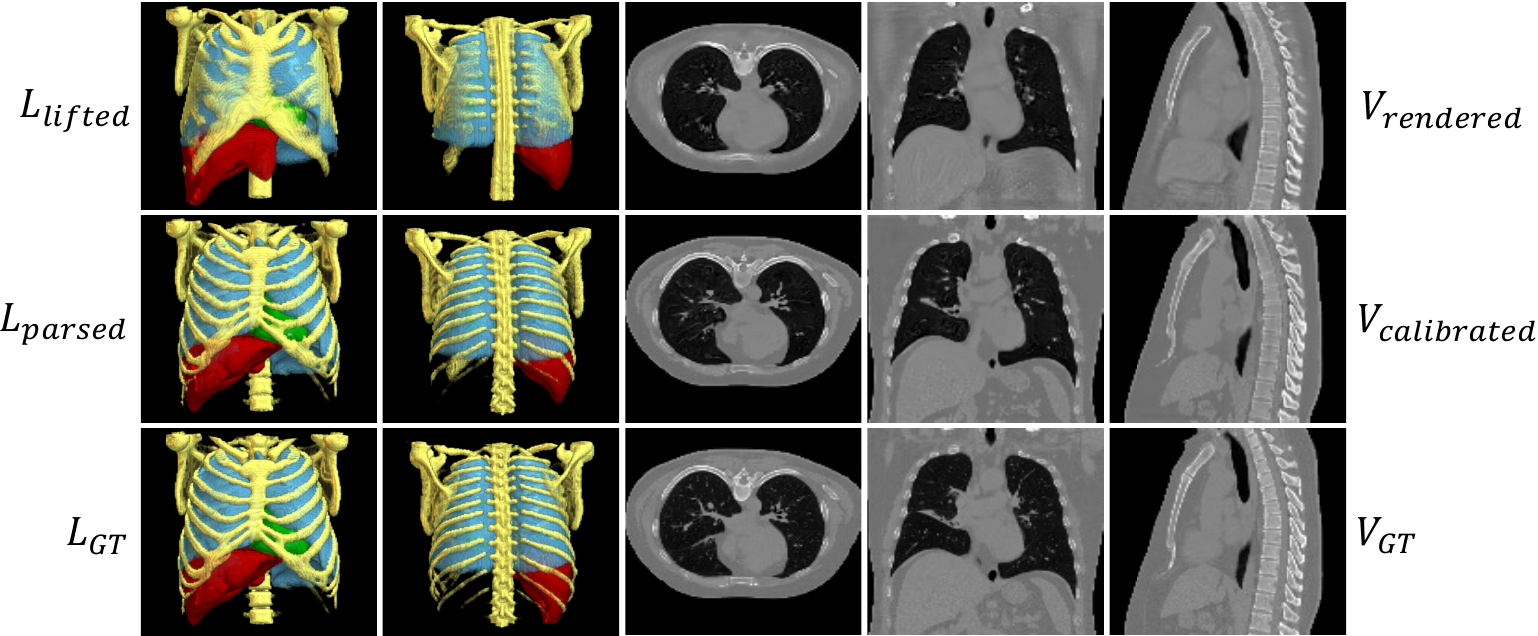}
    \caption{{Intermediate visualization of the layout-intensity refinement process in LiftXR.}
    The first row shows the initial lifted anatomical layout and rendered CT volume, the second row presents the refined layout and calibrated CT after interleaved optimization, and the third row provides the ground-truth reference on the CT-RATE dataset.
    }
    \label{fig:process_visual}
\end{figure}

\paragraph{Analysis of Intermediate Processes:}
As shown in Fig. \ref{fig:process_visual}, 
the visualization further illustrates the effectiveness of the interleaved layout-intensity refinement process in LiftXR. 
Starting from the initial lifted layout and rendered CT volume, the reconstructed CT provides spatially resolved boundary and intensity cues that help refine uncertain anatomical regions. 
Meanwhile, the refined layout introduces more accurate structural constraints for subsequent CT intensity calibration. 
Through this reciprocal interaction, both anatomical layouts and CT volumes are progressively improved, leading to more accurate organ localization, clearer anatomical boundaries, and more consistent intensity distributions compared with the initial reconstruction.

\begin{table}[t]
\centering

\setlength{\tabcolsep}{12pt}
\renewcommand{\arraystretch}{0.8}
\scriptsize

\resizebox{\columnwidth}{!}{%
\begin{tabular}{cccc}
\toprule[1.25pt]
Layout Selection
& Loss Weight $\lambda$ 
& PSNR (dB)$\uparrow$ 
& SSIM (\%)$\uparrow$ \\
\midrule

HU-based Regions
& \multirow{3}{*}{1} 
& 26.07 
& 75.53 \\

Anchor Organs
&  
& \textbf{26.45} 
& \textbf{76.44} \\

All Organs 
& 
& 26.37 
& 76.31 \\

\midrule

\multirow{3}{*}{Anchor Organs}
& 0.1 
& 26.28 
& 75.70 \\

& 1 
& \textbf{26.45} 
& \textbf{76.44} \\

& 10 
& 26.09 
& 75.46 \\

\bottomrule[1.25pt]
\end{tabular}%
}
\caption{{Sensitivity analysis} of anatomical layout selection and the loss weight $\lambda$ on the CT-RATE dataset.}
\label{tab:label}
\end{table}

\subsection{Discussions}
\paragraph{Effect of Layout Selection:}
Tab.~\ref{tab:label} evaluates different anatomical supervision strategies on CT-RATE.
HU-based regions \cite{pan2025draw} provide coarse tissue constraints from intensity ranges, while organ-level supervision introduces explicit anatomical boundaries and spatial relationships.
Both organ anchors and full anatomical labels outperform HU-based regions, demonstrating the importance of organ-level guidance for resolving projection ambiguity.
However, incorporating all possible anatomical targets from TotalSegmentator does not further improve performance since hollow and fine organs may introduce additional segmentation uncertainty, leading to sub-optimal performance.

\paragraph{Effect of $\lambda$:}
Tab.~\ref{tab:label} further investigates the balance between anatomical layout and CT  supervision by varying the loss weight $\lambda$.
The best performance is achieved at $\lambda=1$, suggesting that accurate bi-planar X-ray-to-CT reconstruction requires a joint optimization of anatomical geometry and intensity.
When $\lambda$ is reduced, insufficient anatomical supervision weakens the geometric constraints provided by the predicted layouts, limiting the recovery of organ structures from ambiguous projections.
In contrast, an excessively large $\lambda$ places excessive emphasis on anatomical consistency, which may degrade intensity reconstruction.
Therefore, $\lambda=1$ provides an effective balance between anatomical guidance and CT intensity recovery and is adopted in all experiments.

\begin{figure}[t]
    \centering
    \includegraphics[width=1.0\columnwidth]{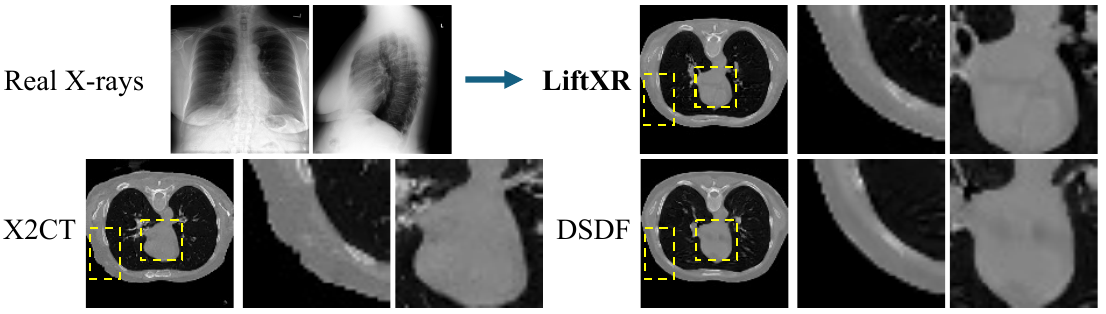} \caption{{Generalization analysis on real frontal and lateral X-rays} on the MIMIC dataset. LiftXR synthesizes more plausible organ textures while reducing boundary distortions.}
    \label{fig:x_real}
\end{figure}

\paragraph{Generalization to Real X-ray Images:}
Fig.~\ref{fig:x_real} shows the results using real frontal and lateral X-rays on the MIMIC dataset \cite{johnson2019mimic}.
Following \cite{ying2019x2ct, kyung2023perspective}, we first train a CycleGAN~\cite{zhu2017unpaired} to transform the style between real X-rays and our synthetic projections, and then reconstruct CT volumes using X2CT, DSDF, and LiftXR.
In the enlarged regions, X2CT produces noticeable boundary artifacts, suggesting inaccurate anatomical recovery under real X-ray inputs.
DSDF preserves more coherent global structures but still fails to recover clear textures.
In contrast, LiftXR produces sharper organ boundaries and more plausible anatomical structures, benefiting from the layout guidance and interleaved layout-intensity refinement.
It demonstrates that LiftXR maintains cross-domain generalization potential on real radiographs.

\begin{table}[t]
\centering

\scriptsize
\setlength{\tabcolsep}{2.5pt}
\renewcommand{\arraystretch}{1.05}

\resizebox{\columnwidth}{!}{
\begin{tabular}{c|cccc|cccc}
\toprule[1.25pt]
\multirow{2}{*}{Method}
& \multicolumn{4}{c|}{[$0^\circ$ , $88^\circ$]}
& \multicolumn{4}{c}{[$0^\circ$ , $92^\circ$]}
\\
\cmidrule(lr){2-5}
\cmidrule(lr){6-9}

& PSNR (dB)$\uparrow$
& $\Delta$
& SSIM (\%)$\uparrow$
& $\Delta$
& PSNR (dB)$\uparrow$
& $\Delta$
& SSIM (\%)$\uparrow$
& $\Delta$
\\
\midrule

X2CT
& 25.49
& 0.47$\downarrow$
& 70.79
& 1.09$\downarrow$
& 25.61
& 0.35$\downarrow$
& 70.92
& 0.96$\downarrow$
\\

DSDF
& 25.20
& 0.62$\downarrow$
& 72.19
& 1.10$\downarrow$
& 25.18
& 0.64$\downarrow$
& 72.12
& 1.17$\downarrow$
\\

\textbf{LiftXR}
& \textbf{26.30}
& \textbf{0.15}$\downarrow$
& \textbf{76.14}
& \textbf{0.30}$\downarrow$
& \textbf{26.27}
&\textbf{0.18}$\downarrow$
& \textbf{75.94}
& \textbf{0.50}$\downarrow$
\\

\bottomrule[1.25pt]
\end{tabular}
}
\caption{{Robustness evaluation under different angle deviations} on the CT-RATE dataset.
The values report the reconstruction performance and the corresponding degradation ($\Delta$) compared with the original acquisition setting.}
\label{tab:robustness}
\end{table}
\paragraph{Robustness to Projection Angle Variations:}
To further validate the robustness of LiftXR under realistic acquisition variations, we investigate the impact of projection angle deviations from the standard orthogonal bi-planar setting.
In clinical scenarios, frontal and lateral X-rays may not be perfectly aligned due to patient positioning and acquisition conditions, introducing geometric inconsistencies between the observed projections and the ideal reconstruction configuration.
As shown in Tab.~\ref{tab:robustness}, we perturb the projection angles to $[0^\circ,88^\circ]$ and $[0^\circ,92^\circ]$ while using the same reconstruction pipeline.
LiftXR consistently achieves the best reconstruction performance under both perturbed settings, while exhibiting the smallest performance degradation compared with existing methods.
This robustness to small projection perturbations is attributed to the proposed interleaved layout-intensity refinement, where anatomical layouts provide explicit geometric constraints and reconstructed CT volumes further refine the estimated layouts.
Such reciprocal interaction enables LiftXR to partially compensate for geometric deviations and maintain reliable volumetric reconstruction under practical and non-ideal projection scenarios.
\paragraph{Limitation and Future Work:}
While LiftXR demonstrates promising performance in bi-planar X-ray-to-CT reconstruction, several limitations remain. 
First, reconstructed CT volumes are currently assessed through image quality and anatomical segmentation metrics, while their effectiveness for clinical applications requires further exploration. 
Second, the interleaved refinement process introduces additional computational costs, which may limit the deployment to specific applications such as mobile scenarios. 
Furthermore, bi-planar X-ray-to-CT reconstruction remains inherently constrained by limited information, making the recovery of highly variable or small pathological anatomical structures challenging. 
Future work will investigate more efficient architectures and conduct broader clinical evaluations to improve the generalizability and applicability of LiftXR.

\section{Conclusion}
We present LiftXR, an interleaved geometry-guided framework for bi-planar X-ray-to-CT reconstruction. LiftXR transitions from projection-conditioned anatomical layout generation to reconstruction-conditioned anatomical perception, allowing the reconstructed volume to refine its geometry and the perceived anatomy to further calibrate region-specific intensities. This generation-perception coupling reduces geometric ambiguity and enables more anatomically consistent volumetric reconstruction. Experiments on two public chest CT datasets demonstrate state-of-the-art reconstruction performance and impressive downstream segmentation results, indicating superior anatomical fidelity.

\bibliography{aaai2027}

@article{jung2021CT,
  title={Basic Physical Principles and Clinical Applications of Computed Tomography},
  author={Jung, Haijo},
  journal={Progress in Medical Physics},
  volume={32},
  number={1},
  pages={1--17},
  year={2021},
  publisher={Korean Society of Medical Physics}
}

@inproceedings{cai2024structure,
  title={Structure-Aware Sparse-View X-ray 3D Reconstruction},
  author={Cai, Yuanhao and Wang, Jiahao and Yuille, Alan and Zhou, Zongwei and Wang, Angtian},
  booktitle={Proceedings of the IEEE/CVF Conference on Computer Vision and Pattern Recognition},
  pages={11174--11183},
  year={2024}
}

@article{lee2018deep,
  title={Deep-Neural-Network-Based Sinogram Synthesis for Sparse-View CT Image Reconstruction},
  author={Lee, Hoyeon and Lee, Jongha and Kim, Hyeongseok and Cho, Byungchul and Cho, Seungryong},
  journal={IEEE Transactions on Radiation and Plasma Medical Sciences},
  volume={3},
  number={2},
  pages={109--119},
  year={2018},
  publisher={IEEE}
}

@article{feldkamp1984FDK,
  title={Practical Cone-beam Algorithm},
  author={Feldkamp, Lee A and Davis, Lloyd C and Kress, James W},
  journal={Journal of the Optical Society of America A},
  volume={1},
  number={6},
  pages={612--619},
  year={1984},
  publisher={Optical Society of America}
}

@article{liu2024geometry,
  title={Geometry-Aware Attenuation Learning for Sparse-View CBCT Reconstruction},
  author={Liu, Zhentao and Fang, Yu and Li, Changjian and Wu, Han and Liu, Yuan and Shen, Dinggang and Cui, Zhiming},
  journal={IEEE Transactions on Medical Imaging},
  volume={44},
  number={2},
  pages={1083--1097},
  year={2024},
  publisher={IEEE}
}

@inproceedings{lin2023learning,
  title={Learning Deep Intensity Field for Extremely Sparse-View CBCT Reconstruction},
  author={Lin, Yiqun and Luo, Zhongjin and Zhao, Wei and Li, Xiaomeng},
  booktitle={International Conference on Medical Image Computing and Computer-Assisted Intervention},
  pages={13--23},
  year={2023},
  organization={Springer}
}

@inproceedings{lin2024learning,
  title={Learning 3D Gaussians for Extremely Sparse-View Cone-Beam CT Reconstruction},
  author={Lin, Yiqun and Wang, Hualiang and Chen, Jixiang and Li, Xiaomeng},
  booktitle={International Conference on Medical Image Computing and Computer-Assisted Intervention},
  pages={425--435},
  year={2024},
  organization={Springer}
}

@inproceedings{ying2019x2ct,
  title={X2CT-GAN: Reconstructing CT From Biplanar X-Rays With Generative Adversarial Networks},
  author={Ying, Xingde and Guo, Heng and Ma, Kai and Wu, Jian and Weng, Zhengxin and Zheng, Yefeng},
  booktitle={Proceedings of the IEEE/CVF Conference on Computer Vision and Pattern Recognition},
  pages={10619--10628},
  year={2019}
}

@inproceedings{kyung2023perspective,
  title={Perspective Projection-Based 3d CT Reconstruction from Biplanar X-Rays},
  author={Kyung, Daeun and Jo, Kyungmin and Choo, Jaegul and Lee, Joonseok and Choi, Edward},
  booktitle={IEEE International Conference on Acoustics, Speech and Signal Processing},
  pages={1--5},
  year={2023},
  organization={IEEE}
}

@inproceedings{liu2024diffux2ct,
  title={DiffuX2CT: Diffusion Learning to Reconstruct CT Images from Biplanar X-Rays},
  author={Liu, Xuhui and Qiao, Zhi and Liu, Runkun and Li, Hong and Zhang, Juan and Zhen, Xiantong and Qian, Zhen and Zhang, Baochang},
  booktitle={European Conference on Computer Vision},
  pages={458--476},
  year={2024},
  organization={Springer}
}

@article{pan2025draw,
  title={Draw Sketch, Draw Flesh: Whole-Body Computed Tomography from Any X-Ray Views},
  author={Pan, Yongsheng and Ye, Yiwen and Zhang, Yanning and Xia, Yong and Shen, Dinggang},
  journal={International Journal of Computer Vision},
  volume={133},
  number={5},
  pages={2505--2526},
  year={2025},
  publisher={Springer}
}

@inproceedings{gopalakrishnan2022fast,
  title={Fast Auto-differentiable Digitally Reconstructed Radiographs for Solving Inverse Problems in Intraoperative Imaging},
  author={Gopalakrishnan, Vivek and Golland, Polina},
  booktitle={Workshop on Clinical Image-Based Procedures},
  pages={1--11},
  year={2022},
  organization={Springer}
}

@article{hamamci2026generalist,
  title={Generalist foundation models from a multimodal dataset for 3D computed tomography},
  author={Hamamci, Ibrahim Ethem and Er, Sezgin and Wang, Chenyu and Almas, Furkan and Simsek, Ayse Gulnihan and Esirgun, Sevval Nil and Dogan, Irem and Durugol, Omer Faruk and Hou, Benjamin and Shit, Suprosanna and others},
  journal={Nature Biomedical Engineering},
  pages={1--19},
  year={2026},
  publisher={Nature Publishing Group UK London}
}

@article{lin2025pixel,
  title={Are Pixel-Wise Metrics Reliable for Sparse-View Computed Tomography Reconstruction?},
  author={Lin, Tianyu and Li, Xinran and Zhuang, Chuntung and Chen, Qi and Cai, Yuanhao and Ding, Kai and Yuille, Alan L and Zhou, Zongwei},
  journal={arXiv preprint arXiv:2506.02093},
  year={2025}
}

@article{shi2026dm4ct,
  title={DM4CT: Benchmarking Diffusion Models for Computed Tomography Reconstruction},
  author={Shi, Jiayang and Pelt, Daniel M and Batenburg, K Joost},
  journal={arXiv preprint arXiv:2602.18589},
  year={2026}
}

@article{guo2024mair,
  title={MAIR-Net: a sparse-view CT reconstruction network based on a combination of mixed attention and iterative optimization learning},
  author={Guo, Tong and Liu, Yi and Zhang, Pengcheng and Liu, Yu and Gui, Zhiguo},
  journal={Journal of Instrumentation},
  volume={19},
  number={08},
  pages={P08029},
  year={2024},
  publisher={IOP Publishing}
}

@article{armato2011lung,
title={The Lung Image Database Consortium (LIDC) and Image Database Resource Initiative (IDRI): A Completed Reference Database of Lung Nodules on CT Scans},
author={Armato III, Samuel G and McLennan, Geoffrey and Bidaut, Luc and McNitt-Gray, Michael F and Meyer, Charles R and Reeves, Anthony P and Zhao, Binsheng and Aberle, Denise R and Henschke, Claudia I and Hoffman, Eric A and others},
journal={Medical Physics},
volume={38},
number={2},
pages={915--931},
year={2011},
publisher={Wiley Online Library}
}

@inproceedings{jeong2025dx2ct,
  title={DX2CT: Diffusion Model for 3D CT Reconstruction from Bi or Mono-planar 2D X-ray (s)},
  author={Jeong, Yun Su and Yoo, Hye Bin and Chun, Il Yong},
  booktitle={International Conference on Acoustics, Speech and Signal Processing},
  pages={1--5},
  year={2025},
  organization={IEEE}
}

@article{jin2017deep,
  title={Deep Convolutional Neural Network for Inverse Problems in Imaging},
  author={Jin, Kyong Hwan and McCann, Michael T and Froustey, Emmanuel and Unser, Michael},
  journal={IEEE Transactions on Image Processing},
  volume={26},
  number={9},
  pages={4509--4522},
  year={2017},
  publisher={IEEE}
}

@inproceedings{dong2019sinogram,
  title={Sinogram interpolation for sparse-view micro-CT with deep learning neural network},
  author={Dong, Xu and Vekhande, Swapnil and Cao, Guohua},
  booktitle={Medical Imaging 2019: Physics of Medical Imaging},
  volume={10948},
  pages={692--698},
  year={2019},
  organization={SPIE}
}

@article{niu2014sparse,
  title={Sparse-view x-ray CT reconstruction via total generalized variation regularization},
  author={Niu, Shanzhou and Gao, Yang and Bian, Zhaoying and Huang, Jing and Chen, Wufan and Yu, Gaohang and Liang, Zhengrong and Ma, Jianhua},
  journal={Physics in Medicine and Biology},
  volume={59},
  number={12},
  pages={2997--3017},
  year={2014},
  publisher={IOP Publishing}
}

@article{goodfellow2014generative,
  title={Generative Adversarial Nets},
  author={Goodfellow, Ian J and Pouget-Abadie, Jean and Mirza, Mehdi and Xu, Bing and Warde-Farley, David and Ozair, Sherjil and Courville, Aaron and Bengio, Yoshua},
  journal={Advances in Neural Information Processing Systems},
  volume={27},
  year={2014}
}

@inproceedings{isola2017image,
  title={Image-To-Image Translation With Conditional Adversarial Networks},
  author={Isola, Phillip and Zhu, Jun-Yan and Zhou, Tinghui and Efros, Alexei A},
  booktitle={Proceedings of the IEEE Conference on Computer Vision and Pattern Recognition},
  pages={1125--1134},
  year={2017}
}

@article{yang2017dagan,
  title={DAGAN: Deep De-Aliasing Generative Adversarial Networks for Fast Compressed Sensing MRI Reconstruction},
  author={Yang, Guang and Yu, Simiao and Dong, Hao and Slabaugh, Greg and Dragotti, Pier Luigi and Ye, Xujiong and Liu, Fangde and Arridge, Simon and Keegan, Jennifer and Guo, Yike and others},
  journal={IEEE Transactions on Medical Imaging},
  volume={37},
  number={6},
  pages={1310--1321},
  year={2017},
  publisher={IEEE}
}

@inproceedings{liu2023solving,
  title={Solving Low-Dose CT Reconstruction via GAN with Local Coherence},
  author={Liu, Wenjie and Ding, Hu},
  booktitle={International Conference on Medical Image Computing and Computer-Assisted Intervention},
  pages={524--534},
  year={2023},
  organization={Springer}
}

@article{armanious2020medgan,
  title={MedGAN: Medical Image Translation using GANs},
  author={Armanious, Karim and Jiang, Chenming and Fischer, Marc and K{\"u}stner, Thomas and Hepp, Tobias and Nikolaou, Konstantin and Gatidis, Sergios and Yang, Bin},
  journal={Computerized Medical Imaging and Graphics},
  volume={79},
  pages={101684},
  year={2020},
  publisher={Elsevier}
}

@article{ho2020denoising,
  title={Denoising Diffusion Probabilistic Models},
  author={Ho, Jonathan and Jain, Ajay and Abbeel, Pieter},
  journal={Advances in Neural Information Processing Systems},
  volume={33},
  pages={6840--6851},
  year={2020}
}

@article{webber2024diffusion,
  title={Diffusion models for medical image reconstruction},
  author={Webber, George and Reader, Andrew J},
  journal={BJR| Artificial Intelligence},
  volume={1},
  number={1},
  pages={ubae013},
  year={2024},
  publisher={Oxford University Press}
}

@article{chung2022score,
  title={Score-based diffusion models for accelerated MRI},
  author={Chung, Hyungjin and Ye, Jong Chul},
  journal={Medical Image Analysis},
  volume={80},
  pages={102479},
  year={2022},
  publisher={Elsevier}
}

@article{wasserthal2023totalsegmentator,
  title={TotalSegmentator: Robust Segmentation of 104 Anatomic Structures in CT Images},
  author={Wasserthal, Jakob and Breit, Hanns-Christian and Meyer, Manfred T and Pradella, Maurice and Hinck, Daniel and Sauter, Alexander W and Heye, Tobias and Boll, Daniel T and Cyriac, Joshy and Yang, Shan and others},
  journal={Radiology: Artificial Intelligence},
  volume={5},
  number={5},
  pages={e230024},
  year={2023},
  publisher={Radiological Society of North America}
}

@inproceedings{zhu2017unpaired,
  title={Unpaired Image-To-Image Translation Using Cycle-Consistent Adversarial Networks},
  author={Zhu, Jun-Yan and Park, Taesung and Isola, Phillip and Efros, Alexei A},
  booktitle={Proceedings of the IEEE International Conference on Computer Vision},
  pages={2223--2232},
  year={2017}
}

@article{johnson2019mimic,
  title={MIMIC-CXR, a de-identified publicly available database of chest radiographs with free-text reports},
  author={Johnson, Alistair EW and Pollard, Tom J and Berkowitz, Seth J and Greenbaum, Nathaniel R and Lungren, Matthew P and Deng, Chih-ying and Mark, Roger G and Horng, Steven},
  journal={Scientific Data},
  volume={6},
  number={1},
  pages={317},
  year={2019},
  publisher={Nature Publishing Group UK London}
}

@inproceedings{wang2017chestx,
  title={ChestX-ray8: Hospital-Scale Chest X-Ray Database and Benchmarks on Weakly-Supervised Classification and Localization of Common Thorax Diseases},
  author={Wang, Xiaosong and Peng, Yifan and Lu, Le and Lu, Zhiyong and Bagheri, Mohammadhadi and Summers, Ronald M},
  booktitle={Proceedings of the IEEE Conference on Computer Vision and Pattern Recognition},
  pages={2097--2106},
  year={2017}
}

@article{shi2025exploration,
  title={Exploration of optimal thresholds for predicting the invasive nature of stage T1 lung adenocarcinoma using artificial intelligence-based 3D solid component volume segmentation},
  author={Shi, Wensong and Hu, Yuzhui and Sun, Yingli and Chang, Guotao and Yang, Yulun and Song, Yinsen and Qian, He and Wei, Zhengpan and Zhao, Liang and Li, Ming and others},
  journal={Quantitative Imaging in Medicine and Surgery},
  volume={15},
  number={1},
  pages={249--258},
  year={2025},
  publisher={LWW}
}

@InProceedings{wu2026codebrain,
    author    = {Wu, Yicheng and Song, Tao and Wu, Zhonghua and Ye, Jin and Ge, Zongyuan and Bai, Wenjia and Chen, Zhaolin and Cai, Jianfei},
    title     = {Virtual Full-stack Scanning of Brain MRI via Imputing Any Quantised Code},
    booktitle = {Proceedings of the IEEE/CVF Conference on Computer Vision and Pattern Recognition},
    month     = {June},
    year      = {2026},
    pages     = {21026-21035}
}

@InProceedings{Wu_2024_CVPR,
     author    = {Wu, Yicheng and Luo, Xiangde and Xu, Zhe and Guo, Xiaoqing and Ju, Lie and Ge, Zongyuan and Liao, Wenjun and Cai, Jianfei},
     title     = {Diversified and Personalized Multi-rater Medical Image Segmentation},
     booktitle = {Proceedings of the IEEE/CVF Conference on Computer Vision and Pattern Recognition},
     month     = {June},
     year      = {2024},
     pages     = {11470-11479}
}

@article{song2026learning,
  title={Learning Modality-Aware Representations: Adaptive Group-wise Interaction Network for Multimodal MRI Synthesis},
  author={Song, Tao and Wu, Yicheng and Hu, Minhao and Luo, Xiangde and Wei, Linda and Wang, Guotai and Guo, Yi and Xu, Feng and Zhang, Shaoting},
  journal={IEEE Transactions on Medical Imaging},
  volume={45},
  number={5},
  pages={2306--2316},
  year={2026},
  publisher={IEEE}
}

@article{lin2026real,
  title={Real-time reconstruction of 3D bone models via very-low-dose protocols},
  author={Lin, Yiqun and Sun, Haoran and Li, Yongqing and Aslam, Rabia and Tse, Lung Fung and Cheng, Tiange and Chui, Chun Sing and Yau, Wing Fung and Le Meur, Victorine R and Amangeldy, Meruyert and others},
  journal={npj Digital Medicine},
  volume={9},
  number={1},
  pages={353},
  year={2026},
  publisher={Nature Publishing Group UK London}
}

@article{pan2026x2shape,
  title={X2Shape: CT-free 3D multi-organ reconstruction with biplanar X-rays},
  author={Pan, Zhaohong and Zhou, Haowei and Ren, Qi and Hou, Xiaorong and Dai, Jingjing and Liu, Xuan and Che, Yongxin and Zhao, Xueqiang and Xie, Yaoqin and Li, Zhicheng and others},
  journal={Medical Image Analysis},
  pages={104085},
  year={2026},
  publisher={Elsevier}
}

@article{yu2025spider,
  title={SPIDER: Structure-Preferential Implicit Deep Network for Biplanar X-ray Reconstruction},
  author={Yu, Tianqi and Tian, Xuanyu and Yang, Jiawen and He, Dongming and Yu, Jingyi and Wang, Xudong and Zhang, Yuyao},
  journal={arXiv preprint arXiv:2507.04684},
  year={2025}
}

@article{chen2025mitigating,
  title={Mitigating Data Consistency Induced Discrepancy in Cascaded Diffusion Models for Sparse-view CT Reconstruction},
  author={Chen, Hanyu and Hao, Zhixiu and Guo, Lin and Xiao, Liying},
  journal={IEEE Transactions on Medical Imaging},
  volume={44},
  number={7},
  pages={3012--3024},
  year={2025},
  publisher={IEEE}
}

@article{jiang2025fast,
  title={Fast-DDPM: Fast Denoising Diffusion Probabilistic Models for Medical Image-to-Image Generation},
  author={Jiang, Hongxu and Imran, Muhammad and Zhang, Teng and Zhou, Yuyin and Liang, Muxuan and Gong, Kuang and Shao, Wei},
  journal={IEEE Journal of Biomedical and Health Informatics},
  volume={29},
  number={10},
  pages={7326--7335},
  year={2025},
  publisher={IEEE}
}

@inproceedings{chen2025cross,
  title={Cross-View Generalized Diffusion Model for Sparse-View CT Reconstruction},
  author={Chen, Jixiang and Lin, Yiqun and Qin, Yi and Wang, Hualiang and Li, Xiaomeng},
  booktitle={International Conference on Medical Image Computing and Computer-Assisted Intervention},
  pages={140--150},
  year={2025},
  organization={Springer}
}

@article{wu2025dual,
  title={Dual-Domain deep prior guided sparse-view CT reconstruction with multi-scale fusion attention},
  author={Wu, Jia and Lin, Jinzhao and Jiang, Xiaoming and Zheng, Wei and Zhong, Lisha and Pang, Yu and Meng, Hongying and Li, Zhangyong},
  journal={Scientific Reports},
  volume={15},
  number={1},
  pages={16894},
  year={2025},
  publisher={Nature Publishing Group UK London}
}

@article{deo2025metrics,
  title={Metrics that matter: Evaluating image quality metrics for medical image generation},
  author={Deo, Yash and Jia, Yan and Lassila, Toni and Smith, William AP and Lawton, Tom and Kang, Siyuan and Frangi, Alejandro F and Habli, Ibrahim},
  journal={arXiv preprint arXiv:2505.07175},
  year={2025}
}

@article{yang2025ct,
  title={CT-SDM: A Sampling Diffusion Model for Sparse-View CT Reconstruction Across Various Sampling Rates},
  author={Yang, Liutao and Huang, Jiahao and Yang, Guang and Zhang, Daoqiang},
  journal={IEEE Transactions on Medical Imaging},
  volume={44},
  number={6},
  pages={2581--2593},
  year={2025},
  publisher={IEEE}
}

@article{chen2018learn,
  title={LEARN: Learned Experts’ Assessment-Based Reconstruction Network for Sparse-Data CT},
  author={Chen, Hu and Zhang, Yi and Chen, Yunjin and Zhang, Junfeng and Zhang, Weihua and Sun, Huaiqiang and Lv, Yang and Liao, Peixi and Zhou, Jiliu and Wang, Ge},
  journal={IEEE Transactions on Medical Imaging},
  volume={37},
  number={6},
  pages={1333--1347},
  year={2018},
  publisher={IEEE}
}

@inproceedings{lin2019dudonet,
  title={DuDoNet: Dual Domain Network for CT Metal Artifact Reduction},
  author={Lin, Wei-An and Liao, Haofu and Peng, Cheng and Sun, Xiaohang and Zhang, Jingdan and Luo, Jiebo and Chellappa, Rama and Zhou, Shaohua Kevin},
  booktitle={Proceedings of the IEEE/CVF Conference on Computer Vision and Pattern Recognition},
  pages={10512--10521},
  year={2019}
}


\end{document}